\pdfoutput=1
\documentclass{article}
\usepackage[utf8]{inputenc}
\usepackage{array}
\usepackage[english]{babel}
\usepackage{xspace}
\usepackage{arxiv}
\usepackage{graphicx}
\usepackage{booktabs}
\usepackage{tcolorbox}
\usepackage{listings}
\usepackage{pgf-pie}
\usepackage{xcolor}
\usepackage{subcaption}
\usepackage{multirow}
\usepackage{amssymb}
\usepackage{amsmath}
\usepackage[T1]{fontenc}    
\usepackage{hyperref}      
\usepackage{url}            
\usepackage{amsfonts}      
\usepackage{microtype}      
\usepackage{graphicx}
\usepackage{natbib}
\usepackage{doi}

\title{Data set creation and Empirical analysis for detecting signs of depression from social media postings}

\author{ {\hspace{1mm}Kayalvizhi S} \\
	Department of Computer Science\\
	SSN College of Engineering\\
	\texttt{kayalvizhis@ssn.edu.in} \\
	\And
     {\hspace{1mm}Thenmozhi D} \\
	Department of Computer Science\\
	SSN College of Engineering\\
	\texttt{theni\_d@ssn.edu.in} \\}

\renewcommand{\shorttitle}{Data set creation and Empirical analysis for detecting signs of depression}
\makeatletter
 \def\@textbottom{\vskip \z@ \@plus 1pt}
 \let\@texttop\relax
\makeatother
\begin{document}
\maketitle

\begin{abstract}
Depression is a common mental illness that has to be detected and treated at an early stage to avoid serious consequences. There are many methods and modalities for detecting depression that involves physical examination of the individual. However, diagnosing mental health using their social media data is more effective as it avoids such physical examinations. Also, people express their emotions well in social media, it is desirable to diagnose their mental health using social media data. Though there are many existing systems that detects mental illness of a person by analysing their social media data, detecting the level of depression is also important for further treatment. Thus, in this research, we developed a gold standard data set that detects the levels of depression as `not depressed', `moderately depressed' and `severely depressed' from the social media postings. Traditional learning algorithms were employed on this data set and an empirical analysis was presented in this paper. Data augmentation technique was applied to overcome the data imbalance. Among the several variations that are implemented, the model with Word2Vec vectorizer and Random Forest classifier on augmented data outperforms the other variations with a score of 0.877 for both accuracy and F1 measure.
\end{abstract}
\section{Introduction}
Depression (major depressive disorder) is a common and serious medical illness that negatively affects the way one feels, thinks and acts \cite{america}. The rate of depression is rapidly increasing day by day. According to Global Health Data Exchange (GHDx), depression has affected 280 million people worldwide  \cite{who}. Detecting depression is important since it has to be observed and treated at an early stage to avoid severe consequences\footnote{https://www.healthline.com/health/depression/effects-on-body}. The depression was generally diagnosed by different methods modalities clinical interviews \cite{al2018detecting}\cite{dibekliouglu2015multimodal}, analysing the behaviour\cite{alghowinem2016multimodal}, monitoring facial and speech modulations\cite{nasir2016multimodal}, physical exams with Depression scales \cite{havigerova2019text}\cite{stankevich2019depression}, videos and audios \cite{morales2016speech}, etc. All these methods of diagnosing involves more involvement of an individual or discussion about their feeling in person. 

On the other hand, social media is highly emerging into our lives with a considerable rate  of increase in social media users according  to the statistics of statista \cite{stat}. Slowly, the social media became a comfortable virtual platform to express our feelings. And so, social media platform can be considered as a source to analyse people's thoughts and so can also be used for analysing mental health of an individual. Thus, we aim to use social media texts for analysing the mental health of a person.\\
The existing works collect social media texts from open source platforms like Reddit \cite{wolohan2018detecting}, Facebook\cite{eichstaedt2018facebook}, Twitter \cite{reece2017forecasting}\cite{tsugawa2015recognizing}\cite{deshpande2017depression}\cite{lin2020sensemood}, Live journals \cite{nguyen2014affective}, blog posts\cite{tyshchenko2018depression}, Instagram \cite{reece2017instagram} etc. and used them to detect depression.\\
\textbf{Research gaps:}\\
All these research works concentrate on diagnosing depression from the social media texts. Although detecting depression has its own significance, detecting the level of depression also has its equal importance for further treatment. Generally, depression is classified into three stages namely mild, moderate and severe \cite{types}. Each stage has its own symptoms and effects and so detecting the level of depression is also a crucial one. Thus, we propose a data set to detect the level of depression in addition to detection of depression from the social media texts . The data set is made available to the public in a CodaLab competition repository  \footnote{\url{https://competitions.codalab.org/competitions/36410}}. This paper explains the process of data set creation that detects the levels of depression along with some baseline models.\\
\textbf{Our contributions in this research include:}
\begin{enumerate}
    \item Creating a new bench mark data set to detect the sign of depression from social media data at postings level.
    \item Developing base line models with traditional learning classifiers.
    \item Analysing the impact of data augmentation 
\end{enumerate}

\section{Related Work}
The aim of our research work is to create a data set that identifies the sign of depression and detect the level of depression and thus, the existing works are analysed in terms of data collection, modalities and methodologies of detecting depression.
\subsection{Modalities and methodologies of depression detection:}
For detecting depression, the data was collected by various methods like clinical interviews \cite{al2018detecting}\cite{dibekliouglu2015multimodal}, analysing the behaviour\cite{alghowinem2016multimodal}, monitoring facial and speech modulations\cite{nasir2016multimodal}, physical exams with Depression scales \cite{havigerova2019text}\cite{stankevich2019depression}, videos and audios \cite{morales2016speech}, etc. Since, the social media users are rapidly increasing day by day, social media data can also be considered as a main source for detecting the mental health. This key idea gave rise to the most utilized data set E-Risk@CLEF-2017 pilot task data set \cite{losada2017erisk} that was collected from Reddit. In addition to this data set, many other data sets such as DAIC corpus \cite{al2018detecting}, AVEC \cite{morales2016speech}, etc. also evolved that detects depression from the social media data. Though few benchmark data set exists to detect depression, more researchers tend to collect data from social media and create their own data sets.

\subsection{Data collection from social media:}
The social media texts were collected from open source platforms like Reddit \cite{wolohan2018detecting}\cite{8681445}, Facebook\cite{eichstaedt2018facebook}, Twitter \cite{reece2017forecasting}\cite{tsugawa2015recognizing}\cite{deshpande2017depression}\cite{lin2020sensemood}, Live journals \cite{nguyen2014affective}, blog posts\cite{tyshchenko2018depression}, Instagram \cite{reece2017instagram} etc. The data from twitter was collected  using API's and annotated into depressed and not depressed classes based on key words like ``depressed, hopeless and suicide" \cite{deshpande2017depression}, using a questionnaire \cite{tsugawa2015recognizing}, survey \cite{reece2017forecasting}, etc. The data was also scrapped from groups of live journals \cite{nguyen2014affective}, blog posts\cite{tyshchenko2018depression} and manually annotated into depressed and not depressed.

Among these social media platforms, Reddit possess large amount text discussion than the other platforms and so Reddit has become widely used platform to collect social media text data recently.

The data were collected from these platforms using Application Programming Interface (API) using hashtags, groups, communities, etc. The data from reddit was collected from Subreddits like ``r/depression help, r/aww, r/AskReddit, r/news, r/Showerthoughts, r/pics, r/gaming, r/depression, r/videos r/todayilearned r/funny" and annotated manually by two annotators into depressed and not depressed class \cite{wolohan2018detecting}. The data was also from subreddits like ``r/anxiety, r/depression and r/depression\_help" and annotated into a data set \cite{pirina-coltekin-2018-identifying}. A data set was  created with classes depression, suicide\_watch, opiates and controlled which was collected using subreddits such as ``r/suicidewatch, r/depression", opioid related forums  and other general forums \cite{yao2020detection}. A survey was also done based on the studies of depression and anxiety from the Reddit data \cite{boettcher2021studies}.

\begin{table}[]
\caption{Comparison of existing data sets} \label{lit}
\centering
\begin{tabular}{lcc}
\toprule
\textbf{Existing system}                             & \textbf{\begin{tabular}[c]{@{}c@{}}Social Media\\  Platform\end{tabular}} & \textbf{Class Labels}                 \\ \toprule
Eichstaedt et.al \cite{eichstaedt2018facebook}       & Facebook &                       Depressed and not depressed               \\ \midrule
Nguyen et.al  \cite{nguyen2014affective}             & Live journal           &                 Depressed and control                     \\ \midrule
Tyshchenko et. al \cite{tyshchenko2018depression}    & Blog post&   Clinical and Control                                   \\ \midrule
Deshpande et.al \cite{deshpande2017depression}       & Twitter  &                        Neutral and negative             \\ \midrule
Lin et.al \cite{lin2020sensemood}                    & Twitter  &                                Depressed and not depressed      \\ \midrule
Reece et.al \cite{reece2017forecasting}              & Twitter  &                                PTSD and Depression      \\ \midrule
Tsugawa et.al \cite{tsugawa2015recognizing}          & Twitter  &   Depressed and not depressed                                   \\ \midrule
Losada et.al \cite{losada2017erisk}                  & Reddit   &                         Depression and Not depression             \\ \midrule
Wolohan et.al \cite{wolohan2018detecting}            & Reddit   &                          Depressed and not depressed            \\ \midrule
Tadesse et.al \cite{8681445}                         & Reddit   &                             Depression indicative and standard         \\ \midrule
Pirina et.al \cite{pirina-coltekin-2018-identifying} & Reddit   &                         positive and negative             \\ \midrule
Yao et.al \cite{yao2020detection}                    & Reddit   &    Depression, Suicide watch, Control and Opiates                                  \\ \midrule
\textbf{Proposed Data set}                           & \textbf{Reddit}  & \begin{tabular}[c]{@{}c@{}}\textbf{Not depressed, moderately depressed} \& \\ \textbf{severely depressed}\end{tabular} \\ \bottomrule
\end{tabular}
\end{table}

From the Table \ref{lit}, it is clear that all these research works have collected the social media data only to detect the presence of depression. Although, diagnosing depression is important, detecting the level of depression is more crucial for further treatment. And thus, we propose a data set that detects the level of depression.
\section{Proposed Work}
We propose to develop a gold standard data set that detects the levels of depression as not depressed, moderately depressed and severely depressed. Initially, the data set was created by collecting the data from the social media platform, Reddit. For collecting the data from archives of Reddit, two way communication is needed, which requires app authentication. After getting proper authentication, the subreddits from which the data must be collected are chosen and the data was extracted. After extracting the data, the data is pre-processed and exported in the required format which forms the data set. The data were then annotated into levels of depression by domain experts following the annotation guidelines. After annotation, the inter-rater agreement is calculated to analyze the quality of data and annotation. Then, the corpus is formed using the mutually annotated instances.Baseline models were also employed on the corpus to analyze the performance. To overcome the data imbalance problem, data augmentation technique was applied and their impact on performance was also analyzed.
\subsection{Data set creation:}
For creating the data set, a suitable social media platform is chosen initially and data is scraped using suitable methods. After scraping the data, the data is processed and dumped in a suitable format.
\subsubsection{Data collection:}
For creating the data set, the data was collected from Reddit \footnote{\url{https://www.reddit.com}}, an open source social media platform since it has more textual data when compared to other social media platforms. This data will be of postings format which includes only one or more statements of an individual. The postings data are scraped from the Reddit archives using the API ``pushshift".
\subsubsection{App authentication:}
For scraping the data from Reddit achieves, Python Reddit API Wrapper(PRAW) is used. The data can be only scraped after getting authentication from the Reddit platform. This authentication process involves creation of an application in their domain, for which a unique client secret key and client id will be assigned. Thus, PRAW allows a two way communication only with these credentials of user\_agent (application name), client\_id and client\_secret to get data from Reddit.
\subsubsection{Subreddit selection}
Reddit is a collection of million groups or forums called subreddits. For collecting the confessions or discussion of people about their mental health, data was scraped from the archives of subreddits groups like ``r/Mental Health, r/depression, r/loneliness, r/stress, r/anxiety".
\subsubsection{Data extraction:}
For each posting, the details such as post ID, title, URL, publish date, name of the subreddit, score of the post and total number of comments can be collected using PRAW. Among these data, PostID, title, text, URL, date and subreddit name are all collected in dictionary format.
\subsubsection{Data pre-processing and exporting:}
After collecting these data, the text and title part are pre-processed by removing the non-ASCII characters and emoticons to get a clean data set. The processed data is exported into a Comma Separated Values (.csv) format file with the five columns. The sample of the collected postings is shown in Table \ref{pdata}.
 
\begin{table}[h]
\caption{Sample Postings data}
%\small
\begin{tabular}{cccccc}
\toprule

\textbf{Post ID} & \textbf{Title} & \textbf{Text}   & \textbf{Url} & \textbf{Publish date} & \textbf{Subreddit} 
\\

\midrule
g69pqt& \begin{tabular}[c]{@{}c@{}}Don’t want \\ to \\ get of bed\end{tabular}  & \begin{tabular}[c]{@{}c@{}}I’m done with me\\  crying all day and\\  thinking to myself \\ that I can’t do a \\ thing and I don’t \\ what to get out \\ of bed at all\end{tabular}& \begin{tabular}[c]{@{}c@{}}https://www.reddit.com\\ /r/depression/\\ comments/g69pqt/ \\ dont\_want\_to\_get\_of\_bed/ \footnote{\url{https://www.reddit.com/r/depression/comments/g69pqt/dont\_want\_to\_get\_of\_bed/}} \end{tabular}    & \begin{tabular}[c]{@{}c@{}}2020-04-23 \\ 02:51:32 \end{tabular}    & depression\\ \midrule
gb9zei& \begin{tabular}[c]{@{}c@{}}Today is a day\\  where \\ I feel emptier\\  than\\  on other days.\end{tabular} & \begin{tabular}[c]{@{}c@{}}It's like I am alone \\ with all my problems. \\ I am sad about the\\  fact I can't trust \\ anyone and\\  nobody could help me\\  because I feel like \\ nobody\\  understand how \\ I feel.  Depression \\ is holding \\ me tight today..\end{tabular} & \begin{tabular}[c]{@{}c@{}}https://www.reddit.com/\\ r/depression/\\ comments/gb9zei/\\ today\_is\_a\_day\_\\ where\_i\_feel\_emptier\\ \_than\_on\_other/  \footnote{\url{https://www.reddit.com/r/depression/comments/gb9zei/today\_is\_a\_day\_where\_i\_feel\_emptier\_than\_on\_other}} \end{tabular} & 
\begin{tabular}[c]{@{}c@{}}2020-05-01 \\ 08:10:06 \end{tabular}  & depression   
\\ \bottomrule
\end{tabular}
\label{pdata}
\end{table}

\subsection{Data Annotation}
After collecting the data, the data were annotated according to the signs of depression. Although all the postings were collected from subreddits that exhibit the characteristics of mental illness, there is a possibility of postings that do not confess or discuss depression. Thus, the collected postings data were annotated by two domain experts into three labels that denote the level of signs of depression namely {``Not depressed, Moderate and Severe"}. Framing the annotation guidelines for postings data is difficult since the mental health of an individual has to be analyzed using his/her single postings. For annotating the data into three classes, the guidelines were formatted as follows:

\subsubsection{Label 1 - Not depressed :} 
The postings data will be annotated as ``Not Depressed", if the postings data reflect one of the following mannerism:
\begin{itemize}
\item If the statements have only one or two lines about irrelevant topics.
    \item If the statements reflect momentary feelings of present situation.
    \item If the statements are about asking questions about any or medication
    \item If the statement is about ask/seek help for friend's difficulties.
\end{itemize}

\begin{tcolorbox}
\textbf{Example 1:}\\

The holidays are the most difficult.\\
Not a big reddit poster, but I felt like this has been past due for myself. The holidays honestly are so hard for me to get through. I've spent the last 6 years of major holidays alone. Mostly because of my retail job, I never get enough time off around the holidays to go home and spend it with family, nor have they been able to visit me. My condolences to anyone else spending this time of year alone no matter what the circumstances may be. 
I moved to a new state 9 months ago and it's been a tough struggle meeting new friends as I didn't know anyone here before I moved.  Now it's new years and all of my "friends" I've made while here yet again flaked on me (was actually excited to have plans for the first time I remember in a while), which I recently found out has been a common occurrence of them just getting together without me. (Which I'm used to at this point, it is what it is). It just sucks knowing you're always the last choice in anyone's lives. And that my depression may be the cause of my `boringness'/lack of interest my friends have towards me. Any tips on making friends for someone struggling mentally? I'm just tired of this constant weight of loneliness bearing down on me. I seriously can't remember the last time someone went out of their way to invite me to something. It seems like I'm always asking to tag along, and then I'm just a burden at that point, which is why I'm starting to lose all hope.\\ 

Whoever takes the time to read this, thank you.
\end{tcolorbox}
\subsubsection{Label 2 - Moderately depressed :}
The postings data will be annotated as ``moderately depressed", if the postings falls under these conditions:
\begin{itemize}
    \item If the statements reflect change in feelings (feeling low for some time and feeling better for some time).
    \item If the statement shows that they aren't feeling completely immersed in any situations
    \item If the statements show that they have hope for life.
\end{itemize}

\begin{tcolorbox}
\textbf{Example 1} :\\

If I disappeared today, would it really matter? \\
I’m just too tired to go on, but at the same time I’m too tired to end it. 
I always thought about this but with the quarantine I just realised it is true. My friends never felt close to me, just like the only two relationships I have ever been in. They never cared about me, to the point where I even asked for help and they just turned a blind eye. 
And my family isn’t any better. 
I don’t know what to do, and I believe it won’t matter if I do something or not.
I’m sorry if my English isn’t good, it isn’t my first language. \\

\end{tcolorbox}

\subsubsection{Label - 3 : Severely depressed :}
The data will be annotated as ``Severely depressed", if the postings have one of the following scenarios:
\begin{itemize}
    \item If the statements express more than one disorder conditions.
    \item If the statements explain about history of suicide attempts.
\end{itemize}

\begin{tcolorbox}
\textbf{Example 1:}\\

Getting depressed again?\\
So I'm 22F and I have taken antidepressants the last time 4 years ago. I've had ups and downs when I got off and with 19 I was having a rough time for two months - started drinking and smoking weed a lot. Kinda managed to get back on track then and haven't been feeling too bad until now. Lately I've been feeling kinda blue and started making mistakes or have to go through stuff multiple times to do it correctly or to be able to remember it. Currently I'm having a week off and have to go back to work on monday. I just don't know I feel like I'm getting worse and want to sleep most of the time and at first I thought it's because I'm used to working a lot, but when I think about having to go back soon I feel like throwing up and at the same time doing nothing also doesn't sit well with me.I guess I'm kinda scared at the moment because I don't want to feel like I was feeling years ago and I still don't feel comfortable with my own mind and don't trust myself that I'm strong enough to pull through if depression hits me again.
\end{tcolorbox}
\subsection{Inter-rater agreement}
After annotating the data, inter-rater agreement was calculated between the decisions of two judges using kappa coefficient estimated using a per-annotator empirical prior over the class labels \cite{artstein2008inter}.
Inter-rater agreement\footnote{https://en.wikipedia.org/wiki/Inter\-rater\_reliability} is the degree of agreement among independent observers who rate, code, or assess the same phenomenon. The inter rater agreement is measured using Cohen's kappa statistics \cite{cohen1960coefficient}. 
\begin{table}[]
\caption{Landis \& Koch measurement table of inter rater agreement}

    \centering
    \begin{tabular}{ll}
    \toprule
    \textbf{Kappa value ($\kappa$)  } &  \textbf{Strength of agreement} \\
    \midrule
$<$ 0          & Poor \\ \midrule
0.01 - 0.20  & Slight \\ \midrule
0.21 - 0.40  & Fair \\  \midrule
0.41 - 0.60  & Moderate  \\  \midrule
0.61 - 0.80  & Substantial  \\ \midrule
0.81 - 0.99  & Almost perfect agreement \\
\bottomrule
    \end{tabular}
    \label{inter}
\end{table}

The inter-rater agreement between the annotations was calculated using sklearn \cite{scikit-learn-a}. For our annotation, the kappa value ($\kappa$) is 0.686. According to Landis \& Koch \cite{landis1977measurement} in the Table \ref{inter}, the $\kappa$ value denotes substantial agreement between the annotators, which proves the consistency of labeling according to the annotation guidelines. Thus, the mutually annotated instances form the corpus.

\subsection{Corpus Analysis}

Initially 20,088 instances of postings data were annotated, out of which 16,613 instances were found to be mutually annotated instances by the two judges, and thus they were considered as instances of data set with their corresponding labels. Table \ref{pdata_analysis} shows the complete statistics of the corpus. 

\begin{table}[]
\caption{Postings data analysis}
\centering
\begin{tabular}{@{}ll@{}}
\toprule
\textbf{Category} & \textbf{Count} \\ \midrule
Total number of instances annotated  & 20,088 \\  \midrule
\begin{tabular}[c]{@{}l@{}} Data set instances \\ (\textit{number of instances mutually annotated)} \end{tabular} & 16,632 \\ \midrule
Total number of sentences  & 1,56,676 \\ \midrule
Total number of words  & 26,59,938 \\ \midrule
Total number of stop-words & 12,47,016 \\ \midrule
Total number of words other than stop-words & 14,12,922 \\ \midrule
Total number of unique words & 28,415 \\ \midrule
Total number of unique stop-words & 150 \\ \midrule
Total number of unique words other than stop-words & 28,265 \\ \midrule
Range of sentences per instance  & 1 - 260  \\ \midrule
Range of words per instance & 1 - 5065  \\ \midrule
Average number of sentences per posting instance  & 9.42 \\ \midrule
Average number of words per posting instance  & 159.92 \\ \bottomrule
\end{tabular}
\label{pdata_analysis}
\end{table}

The whole corpus has 1,56,676 sentences with 26,59,938 words  which shows the size of the corpus created. In the corpus, each posting with its labels is considered as each instance in the corpus. An instance in the corpus will have an average of 9.42 sentences each that varies in the range of 1 to 260 sentences with an average of 159.92 words that lies between 1 to 5065 words.
\begin{figure}[]
    \centering
    \includegraphics[scale=0.5]{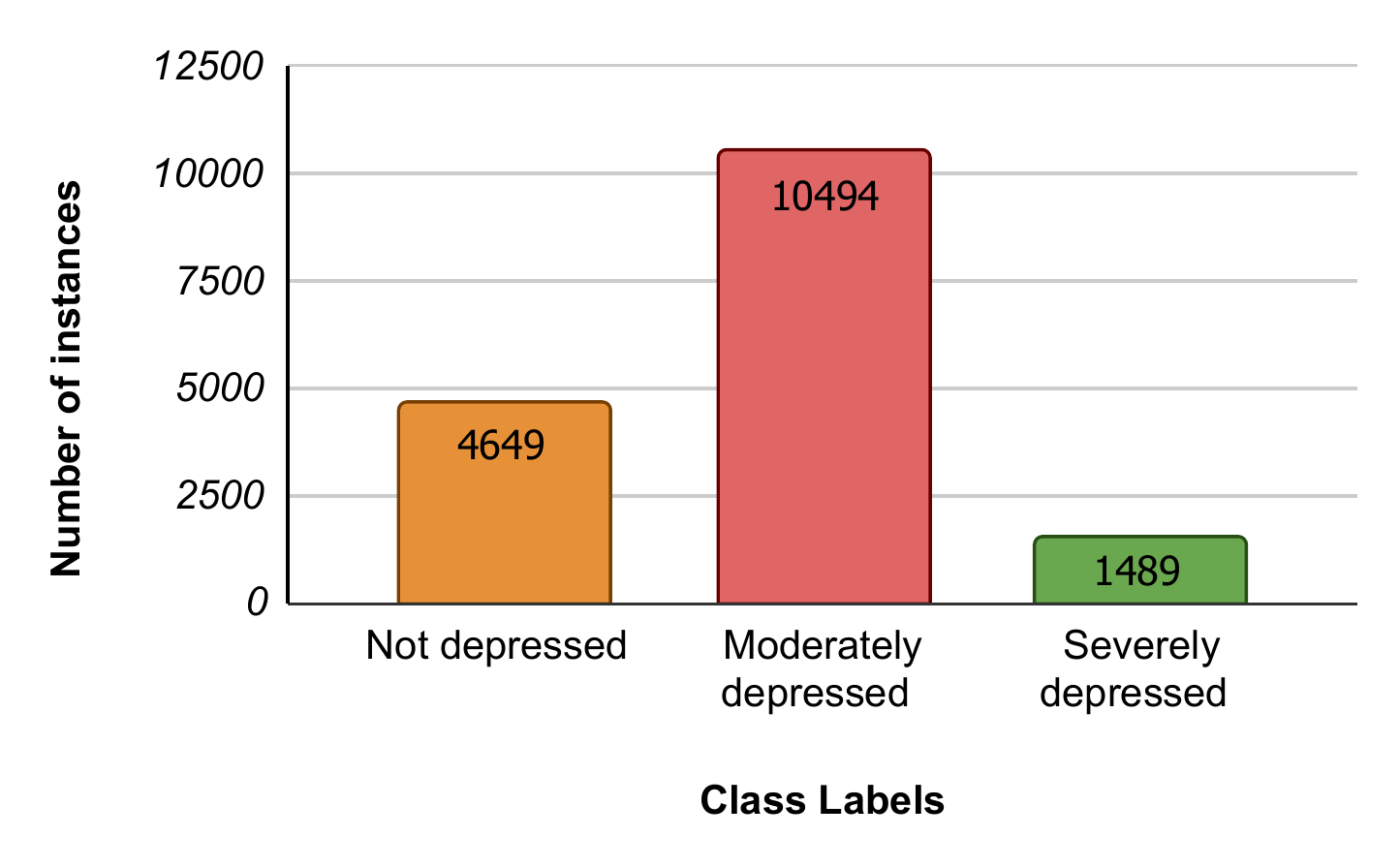}
    \caption{Class wise distribution of the data set}
    \label{label1}
\end{figure}
The distribution of the three class labels in the data set is shown in Figure \ref{label1}. As shown in figure, the data set is unbalanced with 10,494 instances of ``moderately depressed" class, 1489 instances of ``severely depressed" class and 4649 instances of ``Not depressed" class which also includes some duplicate instances.\\

\subsection{Base line models}
The data set has been evaluated using traditional models which are considered as baseline models. The data set has four columns namely id, title, text and class label. For implementation, the title data and text data are initially combined. The combined text data is pre-processed, extracted features, balanced, classified using traditional classifiers and evaluated by cross validation.
\subsubsection{Data Pre-processing:}
The title and text column are combined together as a single text data column by filling the ``NA" instances of both title and text data. The combined text data is cleaned by converting the words to lower case letters and removing unwanted punctuation, ``[removed]" tags, web links, HTML links, stop words and small words(words with length less than two). After cleaning, the instances are tokenized using regexptokenizer \cite{scikit-learn}, stemmed using porter stemmer \cite{porter1980algorithm} and lemmatized using wordnet lemmatizer.

\subsubsection{Feature extraction:} \label{fea}
The features were extracted using three vectorizers namely Word2Vec, Term Frequency - Inverse Document Frequency (TF-IDF) vectorizer and Glove \cite{pennington2014glove} vectorizer. 
\begin{itemize}
    \item \textbf{Word2Vec:} It produces a vector that represents the context of the word considering the occurrence of the word. The vectors are generated using Continuous Bag Of Words.
    \item \textbf{TF-IDF:} It produces a score considering the occurrence of the word in the document. It is based on the relevance of a topic in a particular document. The vectors are calculated using four grams considering a maximum of 2000 features.
    \item \textbf{Glove:} It produces the word embeddings considering the occurrence and co-occurrence of the words with reduced dimensionality. The words are mapped to a word embedding using 6 Billion pre-trained tokens with 100 features each.
\end{itemize}

\subsubsection{Classifiers:} \label{class}
Twelve different classifiers that include Ada Boost Classifier, Decision Tree, Gaussian Naive Bayes, K-Nearest Neighbour, linear-Support Vector Machine, Linear Deterministic Analysis, Logistic Regression,  Multi-layer Perceptron, Qua-\\dratic Deterministic Analysis, Radial Basis Function - Support Vector Machine and Random Forest of Scikit-learn \cite{scikit-learn} were used for classification.  
\begin{itemize}
    \item \textbf{Ada Boost Classifier(ABC):} The Adaptive Boosting algorithm is a collection of N estimator models that assigns higher weights to the mis-classified samples in the next model. In our implementation, 100 estimator models with t{0} random state at a learning rate of {0.1} were used to fine tune the model.
    \item \textbf{Decision Tree (DT):} The decision tree classifier predicts the target value based on the decision rules that was formed using features to identify the target variable. The decision rules are formed using gini index and entropy for information gain. For implementing the decision trees,  the decision  tree classifier was fine tuned with {two} splits of minimum samples of {one} leaf node each by calculating {gini} to choose the {best} split and random state as 0.
    \item \textbf{Gaussian Naive Bayes (GNB):} 
    The Gaussian normal distribution variant of Naive Bayes classifier that depends on the Bayes theorem is Gaussian Naive Bayes. 
    \item \textbf{K-Nearest Neighbour(KNN):} KNN classifies the data point by plotting them and finding the similarity between the data points. In implementation, number of neighbours were set as three with equal weights and euclidean distance as metric to calculate distance.
    \item \textbf{Logistic Regression (LR):} The probabilistic model that predicts the class label based on the sigmoid function for binary classification. As our data set are multi-class data sets, multi-nominal logistic regression was used to evaluate the data sets. For implementation, the classifier was trained with a tolerance of {1e-4}, {1.0} as inverse of regularization strength  and intercept scaling as {1}. 
    \item \textbf{Multi-layer Perceptron (MLP): } The artificial neural network that is trained to predict the class label along with back propagation of error. The multi-layer perceptron of two layers of 100 hidden nodes each was trained with {relu} activation function, {adam} optimizer, learning rate of {0.001} for a maximum {300} iterations. 
    \item \textbf{Discriminant Analysis: } The generative model that utilizes Gaussian distribution for classification by assuming each class has a different co-variance. For implementation, the co-variance is calculated with threshold of {1.0e-04}. Linear DA (LDA) and Quadratic DA (QDA) both were implemented.
    \item \textbf{Support Vector Machine: } The supervised model that projects the data into higher dimensions and then classifies using hyper-planes. The model was trained with {RBF} kernel (RBF-SVM) and linear kernel (L-SVM) function of {three} degree, {0.1} regularization parameter without any specifying any maximum iterations.
    \item \textbf{Random Forest (RF): } Random Forest combines many decision trees as in ensemble method to generate predictions. It overcomes the limitation of decision trees by bagging and bootstrap aggregation. It was implemented with {100} number of estimators.
\end{itemize}

\section{Implementation and Results}
The features extracted in subsection \ref{fea} are classified using the above classifiers in subsection \ref{class} and evaluated using stratified k-fold sampling of Scikit-learn \cite{scikit-learn}. In this validation, data are split into 10 folds and the evaluation results with respect to weighted average F1-score is tabulated in Table \ref{per}.

\begin{table}[]
\caption{F1 score of all baseline models}
\centering
\begin{tabular}{cccc}
\toprule
\textbf{F1 - score} & \textbf{TF- IDF} & \textbf{Glove} & \textbf{Word2Vec} \\ \midrule
\textbf{ABC}        & 0.451& 0.496          & 0.451 \\ \midrule
\textbf{DT}         & 0.469& 0.614          & 0.469 \\ \midrule
\textbf{GNB}        & 0.290& 0.415          & 0.302 \\ \midrule
\textbf{KNN}        & 0.549& 0.604          & 0.594 \\ \midrule
\textbf{L-SVM}      & 0.273& 0.309          & 0.273 \\ \midrule
\textbf{LDA}        & 0.391& 0.395          & 0.391 \\ \midrule
\textbf{LR}         & 0.395& 0.329          & 0.395 \\ \midrule
\textbf{MLP}        & 0.625& \textbf{0.647} & 0.625 \\ \midrule
\textbf{QDA}        & 0.368& 0.459          & 0.368 \\ \midrule
\textbf{RBF -SVM}   & 0.452& 0.560          & 0.452 \\ \midrule
\textbf{RF}         & 0.449& \textbf{0.647} & 0.456 \\ \bottomrule
\end{tabular}
\label{per}
\end{table}

\begin{table}[]
\caption{Accuracy of all baseline models} \label{per1}
\centering
\begin{tabular}{cccc}
\toprule
\textbf{Accuracy} & \textbf{TF- IDF  } & \textbf{Glove  } & \textbf{Word2Vec} \\ \midrule
\textbf{ABC}      & 0.616& 0.654          & 0.616 \\ \midrule
\textbf{DT}       & 0.579& 0.697          & 0.579 \\ \midrule
\textbf{GNB}      & 0.351& 0.464          & 0.351 \\ \midrule
\textbf{KNN}      & 0.695& 0.717          & 0.694 \\ \midrule
\textbf{L-SVM}    & 0.623& 0.646          & 0.623 \\ \midrule
\textbf{LDA}      & 0.619& 0.659          & 0.619 \\ \midrule
\textbf{LR}       & 0.619& 0.650          & 0.619 \\ \midrule
\textbf{MLP}      & 0.700& 0.754          & 0.700 \\ \midrule
\textbf{QDA}      & 0.485& 0.499          & 0.485 \\ \midrule
\textbf{RBF -SVM} & 0.667& 0.733          & 0.667 \\ \midrule
\textbf{RF}       & 0.689& \textbf{0.760} & 0.695 \\ \bottomrule
\end{tabular}
\end{table}
From  the Table \ref{per}, it is clear that the model with Random Forest Classifier and Multi-Layer Perceptron (MLP) applied on the features extracted using Glove performs equally well with an F1-score of 0.647. The performance of the models with accuracy as metric is shown in Table \ref{per1}. From the table, it is clear than the model with Random Forest classifier and Glove vectorizer performs better with an accuracy of 0.760.

\subsection{With Data augmentation}
\begin{figure}[]
\centering
\begin{tabular}{W{c}{0.45\textwidth}W{c}{0.45\textwidth}}
\begin{subfigure}[]{0.44\linewidth}
 \begin{tikzpicture}[scale=0.5,line join=round]
  \pie [polar,color={blue!40, red!30, green!20}]
 {28/,63/, 9/} 
 \end{tikzpicture}
 \caption{Before applying SMOTE}
\end{subfigure}
&
\begin{subfigure}[]{0.44\linewidth}
 \begin{tikzpicture}[scale=0.4]
  \pie [text=legend,color={blue!40, red!30, green!20}]
 {33.3/Not depressed,33.3/Moderately depressed,33.3/Severely depressed} 
 \end{tikzpicture}
 \caption{After applying SMOTE}
\end{subfigure}
\end{tabular}
\caption{Effect of data augmentation}\label{smote}
\end{figure}
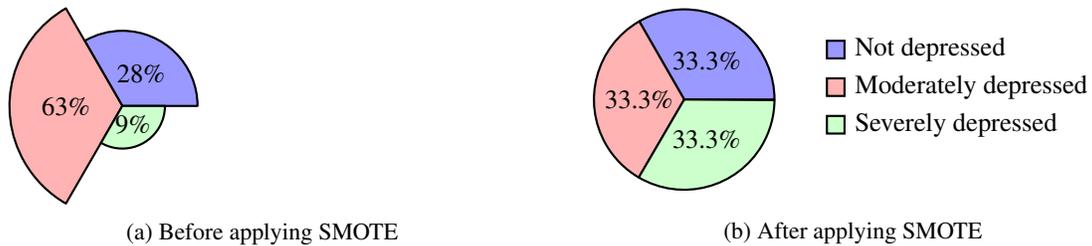

The postings data is populated with more ``moderately depressed" instances and thus, the data has to be balanced before classification for better performance. For balancing the data, Synthetic Minority Oversampling Technique (SMOTE) \cite{Chawla_2002} was applied after vectorization. The effect of augmentation is shown in Figure \ref{smote}. 

The features extracted in subsection \ref{fea} are augmented using SMOTE and then classified using the classifiers in subsection \ref{class}. The performance of these models in terms of F1-score and accuracy after data augmentation are shown in Table \ref{dper} and \ref{dper1} respectively. From the tables, it is clear that the performance was improved and model with Random Forest  classifier applied on the features extracted using Word2Vec performs well with a score of 0.877. 
\begin{table}[]
\caption{F1-score of all baseline models after data augmentation} \label{dper}
\centering
\begin{tabular}{cccc}
\toprule
\textbf{F1 - score} & \textbf{TF- IDF} & \textbf{Glove} & \textbf{Word2Vec} \\  \midrule
\textbf{ABC}        & 0.263& 0.622          & 0.559 \\  \midrule
\textbf{DT}         & 0.273& 0.772          & 0.721 \\  \midrule
\textbf{GNB}        & 0.271& 0.449          & 0.389 \\  \midrule
\textbf{KNN}        & 0.258& 0.814          & 0.834 \\  \midrule
\textbf{L-SVM}      & 0.273& 0.570          & 0.642 \\  \midrule
\textbf{LDA}        & 0.270& 0.550          & 0.540 \\  \midrule
\textbf{LR}         & 0.270& 0.544          & 0.551 \\  \midrule
\textbf{MLP}        & 0.269& 0.775          & 0.852 \\  \midrule
\textbf{QDA}        & 0.276& 0.592          & 0.477 \\  \midrule
\textbf{RBF -SVM}   & 0.273& 0.762          & 0.788 \\  \midrule
\textbf{RF}         & 0.272& 0.854          & \textbf{0.877}    \\ \bottomrule
\end{tabular}
\end{table}
\xspace
\begin{table}[]
\caption{Accuracy of all baseline models after data augmentation}\label{dper1}
\centering
\begin{tabular}{cccc}
\toprule
\textbf{Accuracy} & \textbf{TF- IDF} & \textbf{Glove} & \textbf{Word2Vec} \\ \midrule
\textbf{ABC}      & 0.384& 0.628          & 0.562 \\ \midrule
\textbf{DT}       & 0.388& 0.781          & 0.728 \\ \midrule
\textbf{GNB}      & 0.388& 0.479          & 0.427 \\ \midrule
\textbf{KNN}      & 0.379& 0.839          & 0.854 \\ \midrule
\textbf{L-SVM}    & 0.388& 0.575          & 0.642 \\ \midrule
\textbf{LDA}      & 0.388& 0.550          & 0.550 \\ \midrule
\textbf{LR}       & 0.387& 0.547          & 0.559 \\ \midrule
\textbf{MLP}      & 0.386& 0.780          & 0.857 \\ \midrule
\textbf{QDA}      & 0.393& 0.615          & 0.497 \\ \midrule
\textbf{RBF -SVM} & 0.388& 0.769          & 0.792 \\ \midrule
\textbf{RF}       & 0.388& 0.864 & \textbf{0.877} \\ \bottomrule
\end{tabular}
\end{table}

\section{Research insights}
The researchers can further extend this work by implementing the following methods:
\begin{itemize}
   \item Extend the data set by considering the images along with text data.
    \item Implement deep learning models in the data set.
    \item Implement other methods of data augmentation to improve performance.
\end{itemize}

\section{Conclusions}
Depression is a common mental illness that has to be detected and treated early to avoid serious consequences. Among the other ways of detecting, diagnosing mental health using their social media data seems much more effective since it involves less involvement of the individual. All the existing systems are designed to detect depression from social media texts. Although detecting depression is more important, detecting the level of depression also has its equal significance. Thus, we propose a data set that not only detects depression from social media but also analyzes the level of depression. For creating the data set, the data was collected from subreddits and annotated by domain experts into three levels of depression, namely not depressed, moderately depressed and severely depressed. 
\xspace
An empirical analysis of traditional learning algorithms was also done for evaluating the data sets. Among the models, the model with Glove vectorizer and Random Forest classifier performs well with a F1-score of 0.647 and accuracy of 0.760. 
\xspace
While analyzing the data set, ``the moderately depressed" class seems to be highly populated than the classes and so, a data augmentation method named SMOTE was applied, and the performance is analyzed. Data augmentation improved the performance by 23\% and 12\% in terms of F1-score and accuracy respectively, with both F1-score and accuracy of 0.877.
\xspace
The data set can also be extended by considering the images along with texts for more accurate detection. The work can be extended further by implementing other traditional learning and deep learning models. Other augmentation techniques can also be experimented with for improving the performance of the model.
\section*{Data set availability}
The data set is available to the public in a repository of a Github in the link:
 \url{https://github.com/Kayal-Sampath/detecting-signs-of-depression-from-social-media-postings}.

\bibliographystyle{unsrtnat}
\bibliography{template.bib}
\end{document}